\documentclass[11pt,a4paper]{article}
\usepackage[hyperref]{acl2021}
\usepackage{times}
\usepackage{latexsym}

\usepackage{microtype}

\usepackage{graphicx}

\usepackage{array}

\usepackage[inline]{enumitem}

\aclfinalcopy 

\title{Can vectors read minds better than experts? \\ Comparing data augmentation strategies for the automated scoring of children's mindreading ability}

\author{Venelin Kovatchev\textsuperscript{1}, Phillip Smith\textsuperscript{2}, Mark Lee\textsuperscript{2}, and Rory Devine\textsuperscript{1} \\
  \textsuperscript{1} School of Psychology, University of Birmingham \\
  \textsuperscript{2} School of Computer Science, University of Birmingham \\
  52 Prichatts Road, Edgbaston, B15 2SB, United Kingdom \\
  \{v.o.kovatchev , P.Smith.7 , m.g.lee , R.T.Devine\} @bham.ac.uk}

\date{}

\begin{document}
\maketitle
\begin{abstract}

In this paper we implement and compare 7 different data augmentation strategies for the task of automatic scoring of children's ability to understand others' thoughts, feelings, and desires (or ``mindreading'').

We recruit in-domain experts to re-annotate augmented samples and determine to what extent each strategy preserves the original rating.
We also carry out multiple experiments to measure how much each augmentation strategy improves the performance of automatic scoring systems.
To determine the capabilities of automatic systems to generalize to unseen data, we create UK-MIND-20 - a new corpus of children's performance on tests of mindreading, consisting of 10,320 question-answer pairs. 

We obtain a new state-of-the-art performance on the MIND-CA corpus, improving macro-F1-score by 6 points. 
Results indicate that both the number of training examples and the quality of the augmentation strategies affect the performance of the systems. The task-specific augmentations generally outperform task-agnostic augmentations. Automatic augmentations based on vectors (GloVe, FastText) perform the worst. 

We find that systems trained on MIND-CA generalize well to UK-MIND-20. We demonstrate that data augmentation strategies also improve the performance on unseen data.

\end{abstract}

\section{Introduction}

Many state-of-the-art NLP models are limited by the availability of high quality human-annotated training data. The process of gathering and annotating additional data is often expensive and time consuming. It is especially difficult to gather data for tasks within psychology and psycholinguistics, as test administration typically requires highly trained in-domain experts, controlled environments, and large numbers of human participants.

Data augmentation is a popular technique for artificially enlarging datasets. Typically, data augmentation uses one or more predefined strategies 
to modify existing gold-standard examples while retaining the original label. The objectives of data augmentation are: \begin{enumerate*}[label=\arabic*)] \item to increase the size of the dataset; \item to introduce more variety; \item to reduce overfitting; and \item to improve generalizability. \end{enumerate*}

Data augmentation has been used successfully in computer vision \cite{Shorten2019ASO} and has recently become more popular in the field of NLP \cite{wei-zou-2019-eda,min-etal-2020-syntactic,dai-adel-2020-analysis,10.1007/978-3-030-57321-8_21}. 

We use data augmentation to improve the performance of systems for automatic scoring of children's performance on tests of mindreading (i.e., the ability to reason about others' thoughts, feelings and desires) \cite{HughesDevine2015}. Automatic scoring of mindreading was recently introduced by \citet{kovatchev-etal-2020-mind}. Their corpus, MIND-CA contains hand-scored data from more than 1000 children aged 7 to 14. Collecting data on children's mindreading performance is complicated, time-consuming, and expensive. It requires in-person testing sessions led by trained researchers and children's open-ended responses must be rated by trained annotators. 
Data augmentation could be very beneficial to improve 
the performance and consistency of the automated scoring systems.

In this paper we aim to measure, in a systematic way, the quality and efficiency of the different augmentation strategies. We evaluate and compare the different strategies intrinsically and extrinsically. For the intrinsic evaluation, we recruit in-domain experts to re-annotate augmented examples and determine the extent to which each strategy preserves the original label. 
For the extrinsic evaluation, we measure the quantitative improvement (macro-F1, F1-per-Question, Standard Deviation) of automatic systems on the MIND-CA corpus. Furthermore, we create a new corpus, UK-MIND-20, containing 10,320 question-answer pairs in English and we use it to evaluate the performance of automated systems on unseen data.

We find that the intrinsic ``quality'' of the augmentation strategies varies significantly, according to human raters. However, the extrinsic evaluation demonstrates that all strategies improve the performance of the automated systems. 
We systematically measure the importance of three factors in data augmentation: corpus size, sampling strategy, and augmentation strategy. We find corpus size to be the most important factor. However, the choice of sampling and augmentation strategies also significantly affects the performance of the automated systems. We report a correlation between the ``quality'' of the augmentation and the performance. 
With the best configuration we obtain a new state-of-the-art on MIND-CA, improving Macro-F1 score by 6 points and F1-per-Question by 10.3 points.

We demonstrate that the automated scoring systems can generalize well between MIND-CA and UK-MIND-20. These findings indicate that the methodology for administering and scoring mindreading is consistent and the automatic solutions can be adopted in practice.

The rest of this article is organized as follows. 
Section \ref{maug:rw} discusses the related work. 
Section \ref{maug:method} presents the methodologies for data augmentation. 
Section \ref{maug:quality} compares the quality of the augmentation strategies. 
Section \ref{maug:experiments} describes the machine learning experimental setup and evaluation criteria. 
Section \ref{maug:results} analyzes the effect of data augmentation on automated systems.
Section \ref{maug:dis} presents some follow-up experiments and discusses the implications of the findings. 
Section \ref{maug:fw} concludes the article and proposes directions for future work.

\section{Related Work}\label{maug:rw}

Mindreading (also known as ``theory of mind'') is the ability to understand others' thoughts, feelings, and desires \cite{HughesDevine2015}. For example, in the final scene of Romeo and Juliet, Romeo holds a mistaken belief that Juliet is dead. Being able to understand the state of the world (``Juliet is alive'') and the mistaken belief (``Juliet is dead'') is important to understand the situation and the motivation of the characters.

Individual differences in children's mindreading are linked with both social and academic outcomes and children's wellbeing  \cite{doi:10.1111/j.1467-8624.2011.01669.x,Fink,PMID:26914214}. Furthermore, difficulties with mindreading are linked with a range of mental health problems and neurodevelopmental conditions \cite{COTTER201892}.

The task of automatic scoring of mindreading was first proposed by \citet{kovatchev-etal-2020-mind}. They gathered the responses of 1066 children aged 7-14 on two standardized tests of mindreading: the Strange Story Task \cite{happe} and the Silent Film Task \cite{sfilm}. After digitalizing and manually scoring the responses, they created MIND-CA, a corpus of 11,311 question-answer pairs. They trained and evaluated several automated systems (i.e., SVM, BILSTM, Transformer) and obtained promising initial results. 

Data augmentation is a technique for artificially increasing the size of the dataset. It can also be seen as a type of regularization at the level of the data. Data augmentation can be used to increase the number of instances of specific answer types. It can also introduce more variety, and can reduce the imbalance between classes. Data augmentation is used to improve the performance of automated systems, to reduce the risk of overfitting, and to enhance the ability of automated systems to generalize to unseen data. It is widely used in computer vision \cite{Shorten2019ASO}.

The specifics of natural languages make it more difficult to incorporate data augmentation in NLP. A subtle change to the text can often lead to a substantial difference in meaning and a change of the label. The last two years have seen an increase in the popularity of data augmentation in NLP. \citet{wei-zou-2019-eda} present a Python library that uses simple augmentation methods for improving text classification. \citet{10.1007/978-3-030-57321-8_21} compare different strategies for augmentation in the context of short-text classification. \citet{dai-adel-2020-analysis} compare different data augmentation strategies for the task of Named Entity Recognition. 

Several researchers propose more complex augmentation strategies for NLP. \citet{hou-etal-2018-sequence} propose a sequence-to-sequence model for data augmentation. \citet{kobayashi-2018-contextual} and \citet{gao-etal-2019-soft} use language models in what they call ``contextual augmentation''. \citet{min-etal-2020-syntactic} use syntactic augmentation to improve the performance and generalizability on Natural Language Inference.

In this paper, we take a different approach towards data augmentation. We implement and compare seven different augmentation strategies. Two of the strategies were designed specifically for the task of automatic scoring of mindreading, while the remaining five are task agnostic. We put the emphasis on a systematic evaluation of the augmentation strategies and some key parameters of the augmentation process. We recruit and train in-domain experts to provide intrinsic human evaluation of the data augmentation. We also annotate a new corpus that can measure the performance and improvement on unseen data.

\section{Data Augmentation}\label{maug:method}

We used 7 different strategies for automatic data augmentation. ``Dictionary'' and ``phrase'' strategies make use of task-specific resources, created by in-domain experts. The other 5 strategies (``order'', ``wordnet'', ``ppdb'', ``glove'', ``fasttext'') make use of publicly available task-agnostic resources.

For a source of the augmentation, we used the MIND-CA corpus \cite{kovatchev-etal-2020-mind}. It contains 11,311 question-answer pairs. There are 11 different questions, and an average of 1,028 responses per question. There are three possible labels reflecting the degree to which the response shows context-appropriate mindreading:  0 (fail), 1 (partial score), and 2 (pass). The label distribution for the full corpus is balanced, however the label distribution for the individual questions vary \footnote{For more details, please refer to \citet{kovatchev-etal-2020-mind}}.  

We sought to use data augmentation to create a well-balanced dataset in terms of questions and labels. To achieve this, we created a \textbf{policy for sampling} examples that we used in the augmentation. We split the MIND-CA corpus per question and per label, resulting in 33 question-label sub corpora. The average size of each sub-corpora is 343, and the smallest number of instances in a sub corpora is 160. We sampled 125 examples from each question-label sub-corpus, 375 from each question, for a total 4,125 examples. 

Our sampling strategy ensures that each question-label combination is well represented in the augmentation process. 
In the original MIND-CA corpus, nine question-label pairs had less than 125 instances. As a preliminary step in the data augmentation process, our in-domain experts re-wrote existing responses to improve the balance of the corpus. We used strategy similar to the one used in \citet{hossain-etal-2020-analysis}. We ran statistical and machine learning experiments to ensure that the additional examples do not introduce biases.

For our experiments we initially chose a conservative number of examples (each augmentation increases the original corpus size by 36 \%), to avoid overfitting on the underrepresented question-label pairs. We used a different random state for each augmentation strategy and we ensured that each sample is representative in terms of demographic distribution (age and gender of the participants). 

In a complementary set of experiments, we applied data augmentation directly without the custom sampling strategy. We also experimented with generating larger number of augmented examples (up to 140\% of the original corpus size) via oversampling (see Section \ref{maug:dis}).

In the following subsections, we discuss in more details the different augmentation strategies. 

\subsection{Dictionary Augmentation}

The ``dictionary'' augmentation strategy is a task-specific synonym substitution. We automatically extract the 20 most frequent words for each of the 11 questions, a total of 220 words. We then ask trained corpus annotators to propose a list of synonyms for each word. The synonyms have the same meaning in the context of the particular question. The meaning of the contextual synonyms may not be the same outside of the context. For example, in Silent Film Question \#1, ``men'' can be replaced with ``burglars''. We instruct the experts to create as many synonyms as possible for each word. Some words do not have appropriate contextual synonyms. The final synonym dictionary contains 626 synonyms for 148 words \footnote{The implementation of all augmentation strategies and all resources used (lists of synonyms and introductory phrases) can be found online at \url{https://github.com/venelink/augment-acl21/}}.

The dictionary augmentation algorithm replaces up to two words in each response with their contextual synonyms. The words and their synonyms are selected at random from the available options.

\subsection{Introductory Phrase Augmentation}

The task-specific ``phrase'' augmentation strategy adds a short phrase at the beginning of the response. The appended phrases should not modify the meaning (or score) of the response. An example for such phrase is ``I think (that)''. Our experts create phrases that contain mental state words, such as ``think'', ``know'', and ``believe'', as this category of words is important when scoring children's mindreading ability. Our corpus annotators proposed a list of 15 such phrases. We further modify the 15 phrases with 3 optional conjunctions, resulting in 60 different combinations.
The ``phrase'' augmentation appends a random phrase at the beginning of each response, if the response does not already begin with such a phrase. 

\subsection{Word Replacement Augmentation}

Word replacement augmentation is a strategy that automatically replaces up to two randomly selected words with semantically similar words or phrases. The ``wordnet'' and ``ppdb'' augmentations replace the selected words with a synonym from WordNet \cite{wordnet} or PPDB \cite{pavlickbetter} respectively. The ``glove'' and ``fasttext'' augmentations replace the selected words with the most similar words (or phrases) using  pre-trained GloVe \cite{pennington2014glove} or FastText \cite{joulin2016bag} word embeddings.

We implement the four ``word replacement'' augmentations using the NLP Augmentation python library \cite{ma2019nlpaug}. For this set of experiments we decided not to use BERT-based contextual word embeddings for augmentation, since we are using a DistilBERT classifier. 

\subsection{Change of Order Augmentation}

The ``order'' augmentation strategy changes the position of two words in the sentence. Previous work on data augmentation for NLP \cite{wei-zou-2019-eda,ma2019nlpaug} implement the ``order'' augmentation by changing the position of the two randomly selected words. We enforce a more stringent rule for our algorithm. Specifically, we select one word at random and change its position with one of its neighbouring words. This change is more conservative than picking two words at random. It also reflects the naturally occurring responses from 7- to 14-year-old children in the database. The reorder process is repeated up to two times. 

\subsection{Combining Multiple Augmentations}\label{maug:method:combo}

We also experimented with applying multiple augmentation strategies together. For example the ``dictionary + phrase'' augmentation first replaces up to two words with contextual synonyms and then adds a phrase at the beginning of the response. The data obtained by ``combination'' augmentations was included in the the ``all-lq'' and ``all-hq'' corpora.

\section{Measuring Augmentation Quality}\label{maug:quality}

The quality of data augmentation models in NLP research is typically evaluated extrinsically, by measuring the performance of automated systems trained on augmented data. \citet{wei-zou-2019-eda} propose an intrinsic evaluation 
inspired by the data augmentation research in computer vision. They compare the latent space representations of the original and the augmented sentences and assume that the proximity in latent space indicates that the original labels are conserved.

We argue that a direct comparison of the representation of the texts is not sufficient to determine the quality of the augmentation and the extent to which each strategy preserves the original labels. In natural language, unlike in computer vision, a minor difference in the text and the corresponding representation can cause a significant difference in the meaning of the complex expression and ultimately the label or score assigned to that answer. 

We propose a manual evaluation of the different strategies. For each augmentation strategy, we selected 5 random examples from each question-label sub-corpus, adding up to 165 examples per strategy (4\% of the full sample). Two trained annotators independently rate the augmented pairs for the 7 different augmentation strategies (a total of 1,155 question-answer pairs). To ensure a fair evaluation, the annotators receive a single file with the examples for all augmented strategies shuffled at random. The inter-annotator agreement was 87\% with a Cohen's Kappa of .83.

\begin{table}[h]
    \centering
    \begin{tabular}{|c|c|c|}
    \hline
        \textbf{Augmentation} & \textbf{Quality} & \textbf{Invalid}\\ \hline
        Phrase & 96 & 1\\ 
        Order & 94.5 & 3.5\\
        Dictionary & 94 & 2\\ \hline
        WordNet & 83 & 10 \\
        FastText & 77 & 10 \\
        PPDB & 73 & 12 \\
        GloVe & 68 & 17\\
        \hline
    \end{tabular}
    \caption{Expert comparison of augmentation strategies. 
    Quality - \% of pairs where the label does not change.
    Invalid - \% of pairs where the augmented instance is
    semantically incoherent and cannot be scored.}
    \label{maug:tab:anno}
\end{table}

Table \ref{maug:tab:anno} shows the results of the re-annotation for each augmentation strategy. We define ``quality'' as the \% of examples where the re-annotated label was the same as the original label. We also measure the \% of ``invalid'' examples, where both annotators agreed not to assign a label due to a semantically incoherent response. An example for an incoherent response can be seen in (1).

\begin{itemize}
    \item[(1)] \emph{Why did the men hide ? \\ so telling does'nt get told his .}
\end{itemize}

Based on the analysis, we distinguish between ``high quality'' augmentation strategies (``phrase'', ``order'', and ``dictionary'') and ``low quality'' augmentations (``wordnet'', ``fasttext'', ``ppdb'', and ``glove''). 
The ``high quality'' augmentations preserve the label in over 94\% of the instances and contain less than 4\% invalid responses. 
The ``low quality'' augmentations preserve the label in less than 83\% of the instances and contain more than 10\% invalid responses. According to our raters, GloVe is the worst of all augmentation strategies with 68\% quality and 17\% invalid.

The expert analysis indicates that, at least in our data, there is a substantial difference in the quality of the different augmentation strategies. The task-specific strategies perform much better than the task-agnostic ones, with the exception of ``change of order'' augmentation. In the following sections, we perform a number of machine learning experiments to determine if the quality of the data affects the performance of the automated systems.

\section{Evaluating Automated Systems}\label{maug:experiments}
 
In our experiments, we used the two best systems reported by \citet{kovatchev-etal-2020-mind} - a BiLSTM neural network and a DistilBERT transformer. These systems obtained good results on the original MIND-CA corpus and at the same time were lightweight enough to be implemented in a practical end-to-end application for automatic scoring. We used the same configuration and hyperparameters as reported by \citet{kovatchev-etal-2020-mind}. We modified the existing classes to incorporate and keep track of data augmentation and to implement additional evaluation on UK-MIND-20. All of our code and data are available online \footnote{\url{https://github.com/venelink/augment-acl21/}}.
 
\subsection{Automated Systems. Training setup.}\label{maug:exp:tr}

\begin{table}[h]
    \centering
    \begin{tabular}{|c|c|m{4cm}|}
    \hline
        \textbf{Corpus} & \textbf{Size} & \textbf{Corpus Contents} \\ \hline
        orig & 11,311 & The MIND-CA corpus \\ \hline
        uk-20 & 10,320 & The UK-MIND-20 corpus \\ \hline \hline
        phrase & 15,436 &  MIND-CA + ``phrase'' \\ \hline 
        dict & 15,436 & MIND-CA + ``dictionary'' \\ \hline
        order & 15,436 & MIND-CA + ``order''  \\ \hline
        wordnet & 15,436 & MIND-CA + ``wordnet'' \\ \hline
        fasttext & 15,436 & MIND-CA + ``fasttext''  \\ \hline
        ppdb & 15,436 &  MIND-CA + ``ppdb'' \\ \hline
        glove & 15,436 &  MIND-CA + ``glove'' \\ \hline \hline
        ab-lq & 27,811 &  MIND-CA + ``wordnet'', ``fasttext'', ``ppdb'', and ``glove'' \\ \hline
        all-lq & 44,311 & MIND-CA + ``wordnet'', ``fasttext'', ``ppdb'', and ``glove'' + all 4 synonym substitutions combined with reorder \\ \hline
        ab-hq &  23,686 & MIND-CA + ``phrase'', ``dictionary'' and ``order'' \\ \hline 
        all-hq & 40,186 & MIND-CA + ``phrase'', ``dictionary'', and ``order'' + all four possible combinations of the three strategies \\ \hline
    \end{tabular}
    \caption{All Augmented Training Sets}
    \label{maug:tab:corp}
\end{table}

We trained each of the automated systems on 13 different training sets, shown in Table \ref{maug:tab:corp}. Each set includes the original corpus (MIND-CA) and a number of augmented samples. For example, the \textbf{phrase} dataset contained the 11,311 examples from MIND-CA + 4,125 from the ``phrase'' augmentation, for a total of 15,436 examples. 

In addition to the 7 ``basic'' augmented training sets (one for each augmentation strategy), we also created 4 larger training sets, containing augmented samples from multiple different strategies. The ``All Bassic HQ'' (\textbf{ab-hq}) dataset contains the 11,311 examples from MIND-CA + 4,125 from ``phrase'' + 4,125 from ``dictionary'' + 4,125 from ``order'' for a total of 23,686 examples. Similarly, the ``All Basic LQ'' (\textbf{ab-lq}) dataset contains 27,811 examples from MIND-CA + ``wordnet'', ``fasttext'', ``ppdb'', and ``glove''. 

The two largest datasets, the \textbf{all-lq} and the \textbf{all-hq} datasets contain the corresponding ``all basic'' datasets and additional examples obtained by consecutively applying more than one augmentation strategy to the same original data (the ``combined'' augmentations described in Section \ref{maug:method:combo}). We kept the ``low quality'' and the ``high quality'' data separated, so we can measure the correlation between the ``quality'' and the performance of the automated systems.

\subsection{The UK-MIND-20 Corpus}\label{maug:exp:uk20}

One of the objectives behind data augmentation is to improve the capabilities of automated systems to generalize to unseen data. However, finding unseen data for the same task is often non-trivial, so researchers typically use train-test split or 10-fold cross validation to evaluate the models.

To provide a fair evaluation benchmark for generalizability, we created a new corpus of children's mindreading ability, the UK-MIND-20 corpus. The data for the corpus is part of our own research on children's mindreading in large-scale study involving 1020 8- to 13-year-old children (556 girls, 453 boys, 11 not disclosed) from the United Kingdom. Children completed three mindreading tasks during whole-class testing sessions led by trained research assistants: Strange Stories task \cite{happe}, Silent Film task \cite{sfilm}, and Triangles Task \cite{triangles}. 

Each child answered 14 questions: five from the Strange Story Task, six from the Silent Film Task, and three from the Triangles Task. We do not use the responses for the Triangles task for the evaluation of data augmentation, since that task is not part of the MIND-CA corpus. We obtained a total of 10,320 question-answer pairs for the Strange Stories and the Silent Film portion of the corpus.
Similar to MIND-CA, UK-MIND-20 also includes the age and gender of the participants and responses to a standardized verbal ability test \cite{rust-2008}.  

The children's responses were scored by two trained research assistants, the same assistants that
measured the augmentation quality in Section \ref{maug:quality}. Each response was scored by one annotator. The inter-annotator agreement was measured on a held-out set of questions. We report an inter-annotator agreement of 94\% and a Fleiss Kappa score of .91.

When creating UK-MIND-20, we used the same procedures for administering, scoring, and digitalizing the children responses as the ones used by \citet{kovatchev-etal-2020-mind}. The data for the UK-MIND-20 corpus is gathered in a different time-frame (Oct 2019 -- Feb 2020) and from different locations than MIND-CA (2014 -- 2019).

\subsection{Evaluation Criteria}\label{maug:exp:ev}

The task defined by \citet{kovatchev-etal-2020-mind} consists of scoring the children's mindreading abilities based on the open-text responses to 11 different questions from the Strange Stories Task and the Silent Film Task using three categories (i.e., fail, partial, pass). A single automated system has to score all 11 questions. In this paper we evaluate the system performance in three ways:

\par \textbf{Overall F1}: The macro-F1 on the full test set, containing all 11 questions, shuffled at random.
\par \textbf{F1-per-Q}: We split the test set on 11 parts, one for each question. We obtain the macro-F1 score on each question and calculate the average.
\par \textbf{STD-per-Q}: Similar to F1-per-Q, we obtain the macro-F1 for each question and then calculate the standard deviation of the performance per question.

The \textbf{Overall F1} measures the performance of the system on the full task. \textbf{F1-per-Q} and \textbf{STD-per-Q} measure the consistency of the system across the different questions. 
A practical end-to-end system needs to obtain good results in both.
The additional data facilitates the statistical analysis of the system performance. This evaluation methodology was proposed by \citet{kovatchev-etal-2019-qualitative}.

For each system we performed a 10-fold cross validation using each corpus from Table \ref{maug:tab:corp}. For each fold, we evaluated on both the corresponding test set
and on the full UK-MIND-20 corpus. Our code dynamically removes from the current training set any augmented examples that are based on the current test set to ensure a fair evaluation. All test sets contain only gold-standard human-labeled examples and do not include any augmented data.

\section{Results}\label{maug:results}

\begin{table*}[t]
    \centering
    \begin{tabular}{|c||c|c|c||c|c|c|}
        \hline
        \textbf{Training Set} & \textbf{Test-F1} & \textbf{Test-F1-Q} & \textbf{Test-STD} & \textbf{UK20-F1} & \textbf{UK20-F1-Q} & \textbf{UK20-STD} \\
        \hline
        Orig (baseline)     & .925 & .877 & .059 & .889 & .839 & .063 \\ \hline
        UK-MIND-20        & .893 & .844 & .058 & .890 & .839 & .063 \\ \hline
        Phrase      & .946 & .930 & .031 & \textbf{.893} & \textbf{.854} & .024 \\ 
        Dictionary  & \textbf{.947} & \textbf{.936} & .028 & .892 & .853 & .024 \\ 
        Order       & .947 & .933 & \textbf{.025} & .891 & .852 & \textbf{.022} \\ \hline
        FastText    & .942 & .924 & .030 & .890 & .851 & .023 \\
        GloVe       & .942 & .925 & \textbf{.028} & .891 & .849 & \textbf{.021} \\
        PPDB        & .946 & .929 & .030 & .893 & .851 & .022 \\
        WordNet     & \textbf{.947} & \textbf{.932} & .033 & \textbf{.894} & \textbf{.853} & .023 \\ \hline
        AB-LQ    & .967 & .957 & .021 & .895 & .855 & .021 \\
        AB-HQ    & .972 & .963 & .022 & .897 & .858 & \textbf{.020} \\
        All-LQ          & .978 & .973 & .015 & .895 & .957 & .021 \\
        All-HQ          & \textbf{.985} & \textbf{.980} & \textbf{.011} & \textbf{.898} & \textbf{.858} & .023 \\ \hline
        
    \end{tabular}
    \caption{Performance of a DistilBERT classifier using different augmented sets for training. We report F1, F1-per-Question and standard deviation (per question) on two corpora: Test (MIND-CA), and UK20 (UK-MIND-20).}
    \label{maug:tab:res}
\end{table*}

Table \ref{maug:tab:res} presents the results of the 13 different training configurations with the DistilBERT transformer, using both question and answer as input\footnote{We carried out 4 different sets of experiments: two classifiers (BILSTM and DistilBERT) and two different input setups (i.e., only the answer or both question and answer). Due to space restrictions, we report only the results for the best system, DistilBERT (question + answer). The findings apply to all sets of experiments. The code and results for all experiments are available online at \url{https://github.com/venelink/augment-acl21/} }. The numbers are the average across 10-fold cross validation. For reference, we also include the results obtained by training the system on UK-MIND-20 and testing on MIND-CA. 

The DistilBERT architecture is the best performing system from \newcite{kovatchev-etal-2020-mind}. The baseline system, trained on the original data already obtained very good results: .925 F1 and .877 F1-per-Q on the MIND-CA corpus and .889 F1 and .839 F1-per-Q on the UK-MIND-20 corpus. We demonstrate that systems trained on either of the two datasets can generalize well on the other one (F1 of .89 and F1-per-Q of .84). 
This indicates that the two corpora are compatible and that automated systems can generalize to unseen data. 

It is evident in the table that all of the augmentation strategies successfully improved the performance of the automated systems across all evaluation criteria. For the MIND-CA corpus: F1 improved between 1.7 points (FastText) and 6 points (All-HQ); F1-per-Qiestion improved between 4.7 points (FastText) and 10.3 points (All-HQ); STD-per-Question was reduced by between 1.6 points (WordNet) and 4.8 points (All-HQ). For the UK-MIND-20 corpus: F1 improved between 0.1 point (FastText) and 0.9 point (All-HQ); F1-per-Question improved between 1 point (GloVe) and 1.9 points (All-HQ); STD-per-Question was reduced between 3.9 points (dictionary) and 4.2 points (AB-HQ).

Based on these results, we can draw two conclusions.
\textbf{First}, data augmentation can successfully be used to improve the performance of the systems on the MIND-CA corpus. \textbf{Second}, data augmentation also improves the performance of the automated systems on the unseen examples from UK-MIND-20. While the improvement is not as substantial as seen on MIND-CA, the improvement on all three criteria 
on UK-MIND-20 indicates that the systems are not just overfitting to MIND-CA.

We use the Autorank Python library \cite{Herbold2020} to carry out a statistical analysis on the results and compare the performance gain from each of the augmentation strategies. We use the data from both algorithms and input formats, a total of 480 machine learning models, 40 for each dataset. Based on the provided data, Autorank determines that the most appropriate statistical test is the Friedman-Nemeyni test \cite{Demsar}. The Friedman test reports that there is a statistically significant difference between the median values of the populations. That means that some training sets are consistently performing better (or worse) than others. The post-hoc Nemenyi test can be used to determine and visualise which training sets are better and which are worse.  

\begin{figure}[h!]
\begin{center}
\includegraphics[width=7.2cm]{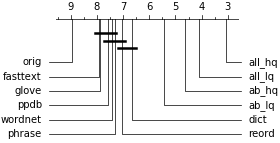}
\end{center}
\caption{Critical Difference Diagram (all). \\ 
Average ranking of training sets (lower is better). \\ 
Connected with a line =\textgreater not statistically significant.}
\label{maug:fig:all}
\end{figure}

Figure \ref{maug:fig:all} shows the Critical Difference diagram of the post-hoc Nemenyi test for all training sets. Each set is plotted with its average ranking across all systems. The difference between systems connected with a line is not statistically significant. The original corpus is the worst performing of all datasets with an average rank of 9. The 7 ``basic'' training sets are grouped in the middle (rank 6.5 to 8). That is, they are all better than the original corpus, but worse than the combined training sets. There is a significant difference between ``All-HQ'', ``All-LQ'', ``AB-HQ'', and ``AB-LQ''. Collectively they are also better than the original training set and the ``basic'' training sets.  

\begin{figure}[h!]
\begin{center}
    
\includegraphics[width=7.2cm]{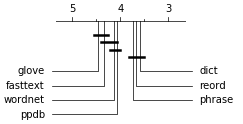}
\end{center}

\caption{Critical Difference Diagram (basic). \\ 
Average ranking of training sets (lower is better). \\ 
Connected with a line =\textgreater not statistically significant.}
\label{maug:fig:basic}
\end{figure}

Figure \ref{maug:fig:basic} shows the Critical Difference diagram of the post-hoc Nemenyi test applied only to the 7 ``basic'' augmentations. After removing the outliers (the original corpus and the collections of multiple augmentation), we can observe a clear, statistically significant distinction between ``high quality" augmentations (``dictionary'', ``phrase'', and ``order'') and ``low quality" augmentations (``glove'', ``fasttext'', ``wordnet'', and ``ppdb'').  

Based on the statistical analysis, we can draw two additional conclusions.
\textbf{Third}, we found that the most important factor affecting the system performance is the number of training examples. We obtain the best results by combining the examples from various different augmentation strategies. \textbf{Fourth}, we demonstrated that when the training size is comparable, the high quality augmentations improve the performance more than the low quality ones. The difference is significant and is consistent both in ``basic'' datasets and in ``combined'' datasets. Vector based augmentations (GloVe and FastText) are performing worse than augmentations based on task-specific or task-agnostic knowledge bases.

\section{Discussion and Further Experiments}\label{maug:dis}

The intrinsic and extrinsic evaluation presented in Section \ref{maug:quality} and Section \ref{maug:results} answered the main research questions posed in this paper. We demonstrated that data augmentation can improve the performance of automated systems including on novel, unseen data. We found that the data augmentation strategies vary in preserving the original label and in how much they improve the machine learning systems trained on them. We also showed that automated scoring systems can generalize well from MIND-CA corpus to UK-MIND-20 and the other way around. All these findings are important for further research on mindreading. At the same time, our data augmentation strategies and evaluation methodology can also be extended to other tasks and domains, contributing to the research of Data Augmentation in NLP in general.

In this section we present additional experiments and an analysis of the impact of several different factors in the process of data augmentation \footnote{Due to space restrictions, we only discuss the overall tendencies. The actual results are available online.}.

\paragraph{Corpus Size} Our experiments indicated that the most important factor for improving the system performance is the corpus size. In Table \ref{maug:tab:res} the systems that perform best are trained on the largest possible amount of data (all-lq/all-hq). 
To further explore the impact of corpus size, we ran an additional set of experiments. We sampled 500 examples for each question-label subcorpora instead of the original 125, increasing the corpus size four times. For each augmentation strategy this resulted in a corpus approximately the same size as ab-lq. 

As expected, the performance of each system increased with corpus size. The ranking of the individual systems remained similar to the one reported with 125 base examples. ``High quality'' augmentations still performed better than ``low quality'' ones. The F1, F1-per-Q, and STD-per-Q for the ``basic low quality'' strategies was approximately the same as the performance for ab-lq. The F1, F1-per-Q, and STD-per-Q for the ``basic high quality'' strategies was approximately the same as the performance for ab-hq.

This new set of experiments confirmed the importance of corpus size. Even strategies that human experts perceive as ``low quality'' are improving the performance of the automated systems. And while the ranking consistently favors the ``high quality'' augmentations, the absolute difference is relatively small. This is in line with the findings on noisy learning which show that machine learning models can be very noise-tolerant \cite{10.5555/2999611.2999745}. We performed one final experiment by combining the all-lq and all-hq data together, but found no increase or decrease of performance compared with using only the all-hq data.

\paragraph{Sampling Strategy} In our experiments, we designed a sampling strategy to ensure that each question-response combination appears in the training data with sufficient frequency. In a complementary set of experiments, we evaluated the importance of the sampling. For each augmentation strategy, we created an augmented dataset with 1500 examples for each question, using a standard sampling that keeps the original ratio of the responses. The size of the dataset is the same as sampling 500 examples for each of the 3 labels. We found that for all strategies, the sampling improves Test-F1-Q between .6 and 1 point and reduces STD-per-Q by 1 point. This finding validates our choice of sampling strategy.

\paragraph{Augmentation Strategy} In Section \ref{maug:results} we demonstrated that when all parameters (sampling, corpus size) are equal the ``high-quality'' strategies rank higher than the ``low-quality'' ones. While the absolute difference in F1 and STD is relatively small on our datasets, the consistency of the performance of the ``high-quality'' strategies has to be taken into consideration. Furthermore, the quantitative performance is only one factor that has to be considered when choosing a strategy for data augmentation. Reducing the noise in the training data can be a desirable characteristic when interpreting the performance of the neural network models, or when working with sensitive data, such as (e.g.) in the health domain. The task-specific augmentations that we proposed and used may require in-domain experts, however the design is rather simple and the process is not time or labour intensive. After the task-specific resource (dictionary, list of phrases) is created, it can be reused for multiple examples and scales very well with corpus size.

\section{Conclusions and Future Work}\label{maug:fw}

We presented a systematic comparison of multiple data augmentation strategies for the task of automatic scoring of children's mindreading ability. We argued that the nature of natural language requires a more in-depth analysis of the quality and performance of the different data augmentation strategies. We recruited in-domain experts and incorporated them in the process of evaluation.

We demonstrated that, for some of the augmentation strategies (``glove'', ``fasttext'', ``ppdb'') there is a substantial portion of the examples (over 20\%) where the rating changes or cannot be assigned due to semantically incoherent text. These differences in the datasets cannot be captured trivially via the visualisation techniques that are typically used for intrinsic evaluation. We also found that the difference in augmentation quality corresponds to a difference in the performance of automated systems trained on the data. To the best of our knowledge, this is the first evaluation of data augmentation in NLP that involves both expert evaluation and automatic metrics and the first study that demonstrates the connection between the two.

We carried out further experiments measuring the importance of factors such as corpus size and sampling strategy.
Our findings on the quality and efficiency of data augmentation strategies and on the use of task-specific resources are relevant for researchers in the area of data augmentation, specifically in domains where the quality of the training gold examples is important or where the amount of data is very limited.

For the purpose of evaluation, we also created a new corpus: UK-MIND-20. It is  the second corpus for automatic scoring of mind reading in children. We demonstrated that systems trained on MIND-CA generalize well on UK-MIND-20. We also showed that data augmentation improves the performance on unseen data. These findings are promising both for the task of scoring children's mindreading and for the use of data augmentation in NLP. To the best of our knowledge, this is the first work where augmentation is evaluated on novel, unseen data for the same task.

This work opens several directions of future work. As a direct continuation of this research, we will incorporate the best performing automated systems and data augmentation techniques in the work of developmental psychologists. This  will facilitate a large-scale studies on mindreading in children and adolescents. 
We are also exploring the possibility of using NLP to address other time and labour intensive problems within psychology. Open-ended short text responses are widely-used within psychological research and the good results obtained in this paper can be replicated in other similar tasks.

\section*{Acknowledgements}

We would like to thank Imogen Grumley Traynor and Irene Luque Aguilera for the annotation and the creation of the lists of synonyms and phrases. We also want to thank the anonymous reviewers for their feedback and suggestions. This project was funded by a grant from Wellcome to R. T. Devine.

\section*{Ethical Statement}

The study was approved by the University of Birmingham STEM Research Ethics Committee and complies with the British Psychological Society Code of Human Research Ethics (2014). Parents and caregivers were provided with detailed information about the study at least one week in advance of data collection and given the opportunity to opt out of the study. Children were also permitted to opt out of the study on the day of data collection without consequence. Data were anonymous at source as children did not provide names or contact information to the research team.

\bibliography{acl2021}
\bibliographystyle{acl_natbib}

\end{document}